\documentclass[5p,twocolumn,times]{elsarticle}
\usepackage[T1]{fontenc}
\usepackage[utf8]{inputenc}
\usepackage{geometry}
\usepackage{color}
\usepackage{verbatim}
\usepackage{amsmath}
\usepackage{graphicx}

\makeatletter

\providecommand{\tabularnewline}{\\}

\usepackage{color}
\usepackage{graphics}
\usepackage{array}
\usepackage{graphicx}
\usepackage{lastpage}
\usepackage{float}

\usepackage{times}

\usepackage{lineno}

\journal{Computers in Biology and Medicine}

\usepackage{amsmath,amssymb}

\makeatother

\begin{document}
\sloppy
\begin{frontmatter}

\title{Adaptive questionnaires for facilitating patient data entry in clinical
decision support systems: Methods and application to STOPP/START v2}

\author[label1]{Jean-Baptiste Lamy\corref{cor1}}
\ead{jibalamy@free.fr}
\cortext[cor1]{Corresponding author\\This is an author file of a manuscript submitted to Computers in Biology and Medicine ; it is available under Creative Commons Attribution Non-Commercial No Derivatives License. }
\author[label1]{Abdelmalek Mouazer}
\ead{malikmouazer@gmail.com}
\author[label1]{Karima Sedki}
\ead{sedkikarima@yahoo.fr}
\author[label2]{Sophie Dubois}
\ead{sophie.dubois@polesante13.fr}
\author[label2,label3]{Hector Falcoff}
\ead{hector.falcoff@sfr.fr}

\address[label1]{INSERM, Université Sorbonne Paris Nord, Sorbonne Université, Laboratory of Medical Informatics and Knowledge Engineering in e-Health, LIMICS, Paris, France}
\address[label2]{CPTS Paris 13 (Communauté Professionnelle Territoriale de Santé), Paris, France}
\address[label3]{SFTG Recherche (Société de Formation Thérapeutique du Généraliste), Paris, France}
\begin{abstract}
Clinical decision support systems are software tools that help clinicians
to make medical decisions. However, their acceptance by clinicians
is usually rather low. A known problem is that they often require
clinicians to manually enter lots of patient data, which is long and
tedious. Existing solutions, such as the automatic data extraction
from electronic health record, are not fully satisfying, because of
low data quality and availability. In practice, many systems still
include long questionnaire for data entry.

In this paper, we propose an original solution to simplify patient
data entry, using an \emph{adaptive} questionnaire, \emph{i.e.} a
questionnaire that evolves during user interaction, showing or hiding
questions dynamically. Considering a rule-based decision support systems,
we designed methods for translating the system's clinical rules into
display rules that determine the items to show in the questionnaire,
and methods for determining the optimal order of priority among the
items in the questionnaire. We applied this approach to a decision
support system implementing STOPP/START v2, a guideline for managing
polypharmacy. We show that it permits reducing by about two thirds
the number of clinical conditions displayed in the questionnaire.
Presented to clinicians during focus group sessions, the adaptive
questionnaire was found ``pretty easy to use”. In the future, this
approach could be applied to other guidelines, and adapted for data
entry by patients.
\end{abstract}
\end{frontmatter}

\section{Introduction}

Clinical decision support systems (CDSS) \citep{Hak2022} are software
tools aimed at helping clinicians to make medical decisions, typically
regarding diagnosis or therapy. Many CDSS have been proposed for chronic
diseases \citep{Souza-Pereira2020}, most of them implementing the
paper clinical practice guidelines produced by learned societies.
CDSS have the potential for improving healthcare. Meta-analysis \citep{Kwan2020,Bright2012}
showed that CDSS are effective at improving health care processes,
despite their clinical and economic impacts are more difficult to
assess.

However, clinicians acceptance with regard to CDSS is often rather
low, and many reasons have been identified \citep{Khairat2018,Moxey2010,Liberati},
including low computer literacy, low trust, reduction of professional
autonomy, poor integration within the workflow, lack of time and the
need for lots of patient data entry. Here, we will focus on the latter
problem: for running the CDSS and obtaining recommendations, a clinician
has to enter the clinical data of his/her patient, which is a tedious
and time-consuming task. It has been shown that patient data entry
contributes to physician burnout \citep{Collier2017}, but also that
a lot of data entry can be associated with entry errors, leading to
wrong decisions \citep{aidedecision-asti-guide-gbp-evalenligne}.

\begin{comment}
User interface was identified as a ``grand challenge'' for CDSS,
fourteen years ago \citep{aidedecision-difficulte}.
\end{comment}

Several solutions have been proposed for helping with patient data
entry. In particular, semantic interoperability \citep{deMello2022}
permits the CDSS to reuse patient data previously entered elsewhere,
typically in the electronic health record (EHR). For instance, in
France, EHR exist and are usually based on the ICD terminology (International
Classification of Diseases release 10). However, many patient conditions
are still entered in EHR as free-text, and not coded using a medical
terminology, impairing its reuse. This is especially true in primary
care, where EHR are less structured and generalized than at hospital,
and patient data is often spread over several EHR, owned by the GP
and physicians of various medical specialties. Moreover, some CDSS
may require data that is commonly not coded in EHR, either because
it is too specific, or judged too trivial by the clinicians (\emph{e.g.}
all GPs do not enter an ICD10 code for ``constipation''). Another
option, natural language processing \citep{Sandoval2019}, allows
the automatic coding of free-text, but almost never achieve a 100\%
accuracy. Finally, various physicians may use the same medical terminology
in a different manner, resulting to different terms for the same patient
\citep{Alvarez-Estevez2020}. To conclude, none of these solutions
is perfect, and clinicians still have to check the patient data used
by the CDSS and to enter the missing data, if any.

In practice, two approaches are commonly used for patient data entry
in CDSS. The first approach consists of a list of patient conditions
restricted to the conditions specifically relevant for the CDSS, associated
with checkboxes. We used this approach in the past in the ASTI project
\citep{jiba-asti-critique} and in Vidal Reco using icons \citep{Pereira2014}.
However, the list quickly becomes long as the number of parameters
considered by the CDSS increases. The second approach consists of
asking the clinicians to enter all patient conditions, \emph{e.g.}
using a terminology such as ICD10. But this is even more tedious.

In various domains, adaptive questionnaires have been used to facilitate
data entry. An adaptive questionnaire is a questionnaire that evolves
during user interaction, typically by showing or hiding questions,
based on the previous answers given by the user.

The objective of this work is to propose an adaptive questionnaire
for helping clinicians with patient data entry in rule-based CDSS.
Since drug prescriptions are usually entered and coded in computerized
physician order entry (CPOE), we will focus on the entry of patient
clinical conditions. As a basic example, if a rule triggers if the
patient has both hypertension and type 2 diabetes, an adaptive questionnaire
may hide the ``type 2 diabetes'' question until the ``hypertension''
question is answered positively (or \emph{vice versa}). Our approach
consists of the automatic translation of the CDSS rule base into a
set of rules that determine whether each patient condition is shown,
and the heuristic ordering of patient conditions for finding the best
priority order between them (\emph{e.g.} should we ask for hypertension
or for type 2 diabetes in the example above?). The approach also includes
a dedicated user interface for the questionnaire.

This study takes place in the context of the ABiMed research project
\citep{Mouazer2021_2}, aimed at designing a CDSS for managing polypharmacy
that associates rule-based systems with advanced user interfaces.
Many CDSS have been proposed in that domain \citep{Mouazer2022_revue,Michiels-Corsten2020,Scott2018},
most of them being based on the implementation of guidelines using
patient data manually entered by clinicians or extracted from EHR.
We applied the proposed adaptive questionnaire approach to STOPP/START
v2 \citep{O'Mahony2015}, a guideline for patients with polypharmacy,
\emph{i.e.} consuming five drugs or more. This guideline is commonly
used for performing medication reviews \citep{Beuscart2021}, which
are structured patient interviews carried out by the pharmacist with
the aim of optimizing patient care. In that context, EHR data extraction
is very difficult to consider, because community pharmacists do not
have access to the GP or hospital EHR. In the ABiMed project, we aim
at extracting data from the GP's EHR and sending it to the pharmacist,
however, in France, GP EHR are fragmented over various editors and
formats, and we currently work with a single editor. For GPs using
EHR from another editor, or for verification purpose, it is very important
to propose an optimized user interface for manual data entry for pharmacists.
We evaluated the proposed system in terms of the number of patient
conditions asked, and clinicians opinion during focus group sessions.

The rest of the paper is organized as follows. Section \ref{sec:Related-works}
describes related works on adaptive questionnaires. Section \ref{sec:Methods}
describes the methods used for designing the adaptive questionnaire
and the associated user interface, and its application to STOPP/START
v2. Section \ref{sec:Results} presents the adaptive questionnaire
user interface and the results of the evaluations. Section \ref{sec:Discussion}
discusses the methods and the results, and, finally, section \ref{sec:Conclusion}
concludes.

\section{\label{sec:Related-works}Related works}

Adaptive questionnaires should be distinguished from \emph{dynamic
forms}. A dynamic form is a questionnaire that is dynamically generated
from a knowledge source \citep{Girgensohn1995}. But contrary to an
adaptive questionnaire, a dynamic form does not change during the
user interaction. F Sadki \emph{et al.} \citep{Sadki2018} proposed
an example of dynamic forms, through a web application that generates
a semantically structured web form from an ontology.

Various categories of adaptive questionnaires exist. An adaptive questionnaire
can be \emph{ordered}, when the questions must be answered in a given
order and new questions only appear after the last question answered,
or \emph{unordered}, when questions can be answered in any order and
new questions may appear anywhere in the questionnaire (after or before
the last answered question).

An adaptive questionnaire can be \emph{exhaustive}, when all relevant
questions are shown (\emph{e.g.} pregnancy status is relevant for
a female patient but not for a male patient), or \emph{heuristic},
when only the most important questions are shown, the importance of
the questions being evaluated with regards to their impact on the
decision (\emph{e.g.} pregnancy status may have a low impact on the
decision for a particular female patient, and thus not shown).

Several ordered heuristic adaptive questionnaires were proposed for
assessing learning style in education. E Mwamikazi \emph{et al.} \citep{Mwamikazi2014}
proposed a system that classifies students in Myers-Briggs Type accurately
while asking 81\% less questions than state-of-the-art systems. The
system relies on a question sorting algorithm that takes into account
the discriminative power of the questions with regards to the Myers-Briggs
Type class. A Ortigosa \emph{et al.} \citep{Ortigosa2010} proposed
AH-questionnaire, a system based on the C4.5 algorithm and decision
trees, for reducing the number of questions asked for determining
the Felder-Silverman’s Learning Style Model.

USHER \citep{Chen2010} combines a dynamic form with an ordered heuristic
adaptive questionnaire. The system learns a probabilistic model over
the questions for ordering them. In addition, USHER can also re-ask
questions that have a high probability of being associated with an
erroneous answer.

ADAPQUEST \citep{Bonesana2021} is a Java tool for the development
of ordered heuristic adaptive questionnaires, based on Bayesian networks.
The tool has been applied to the diagnosis of mental disorders.

Several systems have been proposed for helping patients to enter their
medical data. DQueST \citep{Liu2019} proposes an unordered heuristic
adaptive questionnaire for helping patients to find clinical trials
for which they are eligible. The system starts like a free-text search,
as usual search engines. Then, it ranks questions and identify the
most informative ones. The approach is unordered, the user being able
to choose what type of question he/she will answer next (\emph{e.g.}
a question on diagnosis or prescribed treatment). RC Gibson \emph{et
al.} \citep{Gibson2019} proposed an ontology-based ordered exhaustive
adaptive questionnaire in order to help patients with learning disabilities
to report their symptoms. The questionnaire adapts itself according
to the patient answer, but also according to his/her disabilities.
X Kortum \emph{et al.} \citep{Kortum2017} proposed an ordered heuristic
adaptive questionnaire for self-diagnosis. The authors rely on machine
learning for selecting and ordering questions. PC Sherimon \emph{et
al.} \citep{Sherimon2014} proposed an ordered exhaustive adaptive
questionnaire for helping patients to enter medical data. The system
is associated with a CDSS for diabetes.

Chronic-pharma \citep{Villalba-Moreno2022} includes the LESS-CHRON
module, aimed at helping to deprescribe drugs. It proposes an ordered
exhaustive adaptive questionnaire for entering patient drug order
and main conditions. The questionnaire is structured on three levels:
first, the clinician select the anatomic groups of the drugs taken
by the patient (\emph{e.g.} cardiovascular), then the drug classes
(\emph{e.g.} antihypertensives), and finally the prescription characteristics
(\emph{e.g.} treatment duration) and relevant patient conditions (\emph{e.g.}
systolic blood pressure <160 mmHg). At levels 2 and 3, only the items
corresponding to the items selected at the previous level are shown.

\section{\label{sec:Methods}Methods}

In this work, we opted for an \emph{unordered exhaustive} adaptive
questionnaire. Exhaustive, because all relevant questions have to
be asked in a guideline-based CDSS. And unordered, because clinicians
usually expect questions relative to patient conditions to be organized
by anatomic groups, rather than being ordered by importance.

\subsection{Clinical rule formalization}

Let us consider $C=\left\{ C_{1},C_{2},...,C_{n}\right\} $, a set
of $n$ clinical conditions, $D=\left\{ D_{1},D_{2},...\right\} $,
a set of non-clinical conditions (\emph{e.g.} prescribed drugs), and
$R=\left\{ R_{1},R_{2},...,R_{m}\right\} $, a set of $m$ clinical
rules of the following form:

$\begin{array}{cl}
 & C_{1}^{p}\wedge C_{2}^{p}\wedge...\wedge D_{1}^{p}\wedge D_{2}^{p}\wedge...\\
\wedge & \neg C_{1}^{a}\wedge\neg C_{2}^{a}\wedge...\wedge\neg D_{1}^{a}\wedge\neg D_{2}^{a}\wedge...\\
\wedge & \left(C_{1}^{u1}\vee C_{2}^{u1}\vee...\vee D_{1}^{u1}\vee D_{2}^{u1}\vee...\right)\\
\wedge & \left(C_{1}^{u2}\vee C_{2}^{u2}\vee...\vee D_{1}^{u2}\vee D_{2}^{u2}\vee...\right)\wedge...\,\rightarrow A
\end{array}$

where $C^{p}=\left\{ C_{1}^{p},C_{2}^{p},...\right\} $ and $D^{p}=\left\{ D_{1}^{p},D_{2}^{p},...\right\} $
are the clinical and non-clinical conditions that must be present
in the patient for triggering the rule, respectively, $C^{a}=\left\{ C_{1}^{a},C_{2}^{a},...\right\} $
and $D^{a}=\left\{ D_{1}^{a},D_{2}^{a},...\right\} $ are the clinical
and non-clinical conditions that must be absent, $C^{u1}=\left\{ C_{1}^{u1},C_{2}^{u1},...\right\} $
and $D^{u1}=\left\{ D_{1}^{u1},D_{2}^{u1},...\right\} $ are the clinical
and non-clinical conditions that are members of the first union in
the rule, $C^{u2}$ and $D^{u2}$ are those of the second union, \emph{etc}.,
and $A$ is the action triggered by the rule (\emph{e.g.} prescribing
a given drug; actions are not detailed here for the sake of simplicity).
These rules can also be written:

$\begin{array}{cl}
 & \bigwedge\left\{ c\in C^{p}\right\} \\
\wedge & \bigwedge\left\{ d\in D^{p}\right\} \\
\wedge & \bigwedge\left\{ \neg c\in C^{a}\right\} \\
\wedge & \bigwedge\left\{ \neg d\in D^{a}\right\} \\
\wedge & \bigwedge_{z=1}^{\left|U\right|}\left(\bigvee\left\{ c\in C^{uz}\right\} \vee\bigvee\left\{ d\in D^{uz}\right\} \right)\,\rightarrow A
\end{array}$

A clinical rule $R_{i}$ of that form can be formalized as a 6-element
tuple $R_{i}=\left(C^{p},D^{p},C^{a},D^{a},U=\left\{ \left(C^{u1},D^{u1}\right),\left(C^{u2},D^{u2}\right),...\right\} ,A\right)$
where the first four elements are the sets of present and absent clinical
and non-clinical conditions, the fifth element $U$ is the set of
the unions in the rule, each union being formalized as a pair of clinical
and non-clinical conditions, and the sixth element is the action $A$.

For example, here are two clinical rules from STOPP/START:

-~START rule D2: ``Start fibre supplements (\emph{e.g.} bran, ispaghula,
methylcellulose, sterculia) for diverticulosis with a history of constipation''.

-~STOP rule D6: ``Stop antipsychotics (\emph{i.e.} other than quetiapine
or clozapine) in those with parkinsonism or Lewy Body Disease (risk
of severe extra-pyramidal symptoms)''.

They can be formalized as follows:

\noindent $\begin{array}{l}
R_{D2}=constipation\wedge diverticulosis\wedge\neg fibre\rightarrow start(fibre)\end{array}$

\noindent $\begin{array}{l}
{\color{white}R_{D2}}=\left(\begin{array}{l}
C^{p}=\left\{ constipation,diverticulosis\right\} \\
D^{p}=\emptyset\\
C^{a}=\emptyset\\
D^{a}=\left\{ fibre\right\} \\
U=\emptyset\\
A=start(fibre)
\end{array}\right)\end{array}$

\noindent \smallskip{}

\noindent $\begin{array}{l}
R_{D6}=antipsychotics\wedge(parkinsonism\vee Lewy\,\,Body)\\
\,\,\,\,\,\,\,\,\,\,\,\,\,\,\,\,\rightarrow stop(antipsychotics)
\end{array}$

\noindent $\begin{array}{l}
{\color{white}R_{D2}}=\left(\begin{array}{l}
C^{p}=\emptyset\\
D^{p}=\left\{ antipsychotic\right\} \\
C^{a}=\emptyset\\
D^{a}=\emptyset\\
U=\left\{ \left(\left\{ parkinsonism,Lewy\,\,Body\right\} ,\emptyset\right)\right\} \\
A=stop(antipsychotic)
\end{array}\right)\end{array}$

\subsection{Display rule generation}

We call \emph{display rule} a rule that determines whether a given
clinical condition $x$ should be displayed and asked for manual entry
by a clinician. Display rules are distinct from clinical rules, aiming
at diagnosis or therapy. The action of a display rule is the display
of a given clinical condition in the form. The set of conditions that
should be displayed in the form can be obtained by executing all display
rules and showing conditions for which at least one display rule states
that it should be displayed. When the patient profile changes (\emph{e.g.}
a new drug is prescribed, or a clinical condition has been entered),
display rules must be executed again to determine the clinical conditions
to display after the change.

In the design of the algorithm for generating display rules, we made
the following assumptions: (a) It is preferable to display the lowest
possible number of clinical conditions, in order to simplify the questionnaire
and reduce the time required for data entry. (b) Non-clinical conditions
(\emph{e.g.} drug prescriptions or lab test results) are already known
and coded (otherwise, they should be treated as clinical conditions
in the following formula), and thus they will be checked first. (c)
Clinical conditions are more likely to be false than true (this is
verified for almost all conditions, with the possible exception of
hypertension and renal failure for the elderly). Thus, clinical conditions
that must be present will be displayed before those that must be absent.

Consequently, in display rules, the conditions will be considered
in the following order of priority: first, check the $D^{p}$ and
$D^{a}$ member of the rule 6-element tuple, then check $C^{a}$,
finally check $C^{p}$ and $U$. For example, for $R_{D2}$, display
rules will first check than fibre is absent ($D^{a}$), then display
the first clinical condition (constipation or diverticulosis, $C^{p}$),
and display the second one only if the first one is true. Whether
constipation or diverticulosis is shown first is here an arbitrary
choice, thus we have to define an order of priority between clinical
conditions.

Let us consider a strict total order $\prec$ between the clinical
conditions in $C$, such that $c1\prec c2$ means that $c1$ is displayed
preferably than $c2$ when one of them must be chosen to be displayed
first (\emph{e.g.} if a rule has for conditions $c1\wedge c2$, and
if $c1\prec c2$, then $c1$ is displayed, and $c2$ will be displayed
only if $c1$ is present). We will explain later how to find the best
order $\prec$. For each clinical condition $x\in C^{p}$ in rule
$R_{i}$, we define the rule $\mathcal{R}^{p}\left(R_{i},x\right)$
that determines whether present clinical condition $x\in C^{p}$ needs
to be displayed, or not, as follows:

\noindent $\begin{array}{ccl}
\mathcal{R}^{p}\left(R_{i},x\in C^{p}\right) & = & \bigwedge\left\{ c\in C^{p}\mid c\prec x\right\} \\
 & \wedge & \bigwedge\left\{ d\in D^{p}\right\} \\
 & \wedge & \bigwedge\left\{ \neg c\in C^{a}\right\} \\
 & \wedge & \bigwedge\left\{ \neg d\in D^{a}\right\} \\
 & \wedge & \bigwedge_{z=1}^{\left|U\right|}\begin{cases}
\bigvee\left\{ c\in C^{uz}\right\} \vee\bigvee\left\{ d\in D^{uz}\right\} \\
\,\,\,\,\,\,\,\,\,\,\,\,if\,\,\forall c\in C^{uz},c\prec x\\
True\,\,otherwise
\end{cases}\\
 & \rightarrow & display(x)
\end{array}$

$\mathcal{R}^{p}$ displays the condition $x$ if non-clinical conditions
are satisfied ($D^{p}$ and $D^{a}$), absent clinical conditions
are satisfied ($C^{a}$), and present clinical conditions in $C^{p}$
and $U$ that have priority on $x$ are satisfied.

We define two rule $\mathcal{R}^{a1}\left(R_{i},x\in C^{a}\right)$
and $\mathcal{R}^{a2}\left(R_{i},x\in C^{a}\right)$ for each clinical
condition that must be absent:

\noindent $\begin{array}{ccl}
\mathcal{R}^{a1}\left(R_{i},x\in C^{a}\right) & = & \bigwedge\left\{ c\in C^{p}\right\} \\
 & \wedge & \bigwedge\left\{ d\in D^{p}\right\} \\
 & \wedge & \bigwedge\left\{ \neg c\in C^{a}\mid c\neq x\right\} \\
 & \wedge & \bigwedge\left\{ \neg d\in D^{a}\right\} \\
 & \wedge & \bigwedge_{z=1}^{\left|U\right|}\left(\bigvee\left\{ c\in C^{uz}\right\} \vee\bigvee\left\{ d\in D^{uz}\right\} \right)\\
 & \rightarrow & display(x)
\end{array}$

~

\noindent $\begin{array}{ccl}
\mathcal{R}^{a2}\left(R_{i},x\in C^{a}\right) & = & \left\{ x\right\} \\
 & \wedge & \bigwedge\left\{ d\in D^{p}\right\} \\
 & \wedge & \bigwedge\left\{ \neg c\in C^{a}\mid c\prec x\right\} \\
 & \wedge & \bigwedge\left\{ \neg d\in D^{a}\right\} \\
 & \rightarrow & display(x)
\end{array}$

Since clinical conditions are expected to be more frequently false
than true, absent conditions are checked lastly. $\mathcal{R}^{a1}$
displays the condition $x$ if all other conditions are satisfied
(\emph{i.e.} non-clinical conditions $D^{p}$ and $D^{a}$, present
clinical conditions $C^{p}$, unions $U$ and other absent clinical
conditions in $C^{a}$). $\mathcal{R}^{a2}$ displays the condition
$x$ if it is true and if other conditions that have priority are
satisfied. $\mathcal{R}^{a2}$ is needed to permit the user to remove
condition $x$ if it has been set, but the other conditions of the
rules have not been set. This normally cannot occur when interacting
with the system starting from an empty list of clinical conditions;
however, it may occur if the initial list of clinical conditions is
not empty (\emph{e.g.} if some clinical conditions are automatically
extracted from EHR, and then the clinician verifies them and completes
them is needed).

Finally, we define two rules $\mathcal{R}^{u1}\left(R_{i},x\in C^{uk}\right)$
and $\mathcal{R}^{u2}\left(R_{i},x\in C^{uk}\right)$ for each clinical
condition that is a member of a union, with $1\leq k\leq\left|U\right|$.

\noindent $\begin{array}{ccl}
\mathcal{R}^{u1}\left(R_{i},x\in C^{uk}\right) & = & \bigwedge\left\{ c\in C^{p}\mid c\prec x\right\} \\
 & \wedge & \bigwedge\left\{ d\in D^{p}\right\} \\
 & \wedge & \bigwedge\left\{ \neg c\in C^{a}\right\} \\
 & \wedge & \bigwedge\left\{ \neg d\in D^{a}\right\} \\
 & \wedge & \bigwedge_{z=1}^{\left|U\right|}\begin{cases}
\bigwedge\left\{ \neg c\in C^{uz}\right\} \wedge\bigwedge\left\{ \neg d\in D^{uz}\right\} \\
\,\,\,\,\,\,\,\,\,\,\,\,if\,\,x\in C^{uz}\\
\bigvee\left\{ c\in C^{uz}\right\} \vee\bigvee\left\{ d\in D^{uz}\right\} \\
\,\,\,\,\,\,\,\,\,\,\,\,if\,\,\forall c\in C^{uz},c\prec x\\
True\,\,otherwise
\end{cases}\\
 & \rightarrow & display(x)
\end{array}$

~

\noindent $\begin{array}{ccl}
\mathcal{R}^{u2}\left(R_{i},x\in C^{uk}\right) & = & \bigwedge\left\{ c\in C^{p}\mid c\prec x\right\} \\
 & \wedge & \bigwedge\left\{ d\in D^{p}\right\} \\
 & \wedge & \bigwedge\left\{ \neg c\in C^{a}\right\} \\
 & \wedge & \bigwedge\left\{ \neg d\in D^{a}\right\} \\
 & \wedge & \bigwedge_{z=1}^{\left|U\right|}\begin{cases}
\left\{ x\right\} \,\,\,\,if\,\,x\in C^{uz}\\
\bigvee\left\{ c\in C^{uz}\right\} \vee\bigvee\left\{ d\in D^{uz}\right\} \\
\,\,\,\,\,\,\,\,\,\,\,\,if\,\,\forall c\in C^{uz},c\prec x\\
True\,\,otherwise
\end{cases}\\
 & \rightarrow & display(x)
\end{array}$

$\mathcal{R}^{u1}$ displays the condition $x$ if no condition in
the union is true and if other conditions that have priority are satisfied.
$\mathcal{R}^{u2}$ displays the condition $x$ if it true and if
other conditions that have priority are satisfied. Similarly to $\mathcal{R}^{a2}$,
$\mathcal{R}^{u2}$ is only needed to permit the clinician to unset
the condition $x$ if it was set outside of the questionnaire.

Notice that display rules can be expressed as 6-element tuples, exactly
as clinical rules. Considering $R_{i}=\left(C^{p},D^{p},C^{a},D^{a},U,A\right)$,
we have:

\noindent $\mathcal{R}^{p}\left(R_{i},x\in C^{p}\right)=\left(\begin{array}{l}
\left\{ c\in C^{p}\mid c\prec x\right\} \\
D^{p}\\
C^{a}\\
D^{a}\\
\left\{ (C^{u},D^{u})\in U\mid\forall c\in C^{u},c\prec x\right\} \\
display(x)
\end{array}\right)$

~

\noindent $\mathcal{R}^{a1}\left(R_{i},x\in C^{a}\right)=\left(\begin{array}{l}
C^{p}\\
D^{p}\\
C^{a}\setminus\left\{ x\right\} \\
D^{a}\\
U\\
display(x)
\end{array}\right)$

~

\noindent $\mathcal{R}^{a2}\left(R_{i},x\in C^{a}\right)=\left(\begin{array}{l}
\left\{ x\right\} \\
D^{p}\\
\left\{ c\in C^{a}\mid c\prec x\right\} \\
D^{a}\\
\emptyset\\
display(x)
\end{array}\right)$

~

\noindent %
\begin{comment}
\noindent $\mathcal{R}^{a}\left(R_{i},x\in C^{a}\right)=\left(\begin{array}{l}
C^{p}\\
D^{p}\\
C^{a}\setminus\left\{ x\right\} \\
D^{a}\\
\left\{ (C^{u},D^{u})\in U\mid\forall c\in C^{u},c\prec x\right\} \\
display(x)
\end{array}\right)$

\noindent $\mathcal{R}^{u}\left(R_{i},x\in C^{uk}\right)=\left(\begin{array}{l}
\left\{ c\in C^{p}\mid c\prec x\right\} \\
D^{p}\\
C^{a}\cup C^{uk}\setminus\left\{ x\right\} \\
D^{a}\cup D^{uk}\\
\left\{ (C^{u},D^{u})\in U\mid C^{u}\neq C^{uk}\wedge\forall c\in C^{u},c\prec x\right\} \\
display(x)
\end{array}\right)$
\end{comment}

\noindent $\mathcal{R}^{u1}\left(R_{i},x\in C^{uk}\right)=\left(\begin{array}{l}
\left\{ c\in C^{p}\mid c\prec x\right\} \\
D^{p}\\
C^{a}\cup C^{uk}\\
D^{a}\cup D^{uk}\\
\left\{ (C^{u},D^{u})\in U\mid C^{u}\neq C^{uk}\wedge\forall c\in C^{u},c\prec x\right\} \\
display(x)
\end{array}\right)$

~

\noindent $\mathcal{R}^{u2}\left(R_{i},x\in C^{uk}\right)=\left(\begin{array}{l}
\left\{ c\in C^{p}\mid c\prec x\right\} \cup\left\{ x\right\} \\
D^{p}\\
C^{a}\\
D^{a}\\
\left\{ (C^{u},D^{u})\in U\mid C^{u}\neq C^{uk}\wedge\forall c\in C^{u},c\prec x\right\} \\
display(x)
\end{array}\right)$

\begin{comment}
\textbf{A et B et (C ou D)}

~

ordre A < B < C < D :

=> A

A => B

A et B et non D => C

A et B et non C => D

~

ordre A < C < B < D :

=> A

A => B

A et non D => C

A et B et non C => D

~

ordre A < C < D < B :

=> A

A et (C ou D) => B

A et non D => C

A et non C => D

~

~

\textbf{A et B et non C et non D}

~

ordre A < B < C < D :

=> A

A => B

A et B => C

A et B et non C => D

~

ordre A < C < D < B :

=> A

A => B

A et B => C

A et B et non C => D

~

~

\textbf{A et B et non C}

ordre A < B < C :

non C => A

A et non C => B

A et B => C

~

ordre A < C < B :

non C => A

A et non C => B

A et B => C

~

~

\textbf{A et B et (C ou D) et non E}

~

ordre A < B < C < D < E :

non E => A

non E et A => B

non E et A et B => C

non E et A et B => D

A et B et (C ou D) => E

~

ordre A < C < E < B < D :

non E => A

non E et A => B

non E et A => C

non E et A => D

A et B => E

~

ordre A < C < E < B < D :

non E => A

non E et A => B

non E et A => C

non E et A => D

A et B => E
\end{comment}

We denote by $R^{d}$ the set of all display rules generated from
the clinical rules in $R$.

For example, for START rule D2 and STOP rule D6, and considering $constipation\prec diverticulosis$,
we obtain the following display rules:

$\begin{array}{l}
\mathcal{R}^{p}\left(R_{D2},constipation\right)=\neg fibre\\
\,\,\,\,\,\,\,\,\,\,\,\,\,\,\,\,\,\,\,\,\,\,\,\,\,\,\,\,\,\,\,\,\,\,\,\,\,\,\,\,\,\,\,\,\,\,\,\,\,\,\,\,\,\,\,\,\,\,\,\,\,\rightarrow display(constipation)
\end{array}$

$\begin{array}{l}
\mathcal{R}^{p}\left(R_{D2},diverticulosis\right)=constipation\wedge\neg fibre\\
\,\,\,\,\,\,\,\,\,\,\,\,\,\,\,\,\,\,\,\,\,\,\,\,\,\,\,\,\,\,\,\,\,\,\,\,\,\,\,\,\,\,\,\,\,\,\,\,\,\,\,\,\,\,\,\,\,\,\,\,\,\,\,\,\rightarrow display(diverticulosis)
\end{array}$

\smallskip{}

$\begin{array}{l}
\mathcal{R}^{u1}\left(R_{D6},parkinsonism\right)=antipsychotics\wedge\neg Lewy\,\,Body\\
\,\,\,\,\,\,\,\,\,\,\,\,\,\,\,\,\,\,\,\,\,\,\,\,\,\,\,\,\,\,\,\,\,\,\,\,\,\,\,\,\,\,\,\,\,\,\,\,\,\,\,\,\,\,\,\,\,\,\,\,\,\,\,\,\rightarrow display(parkinsonism)
\end{array}$

$\begin{array}{l}
\mathcal{R}^{u2}\left(R_{D6},parkinsonism\right)=antipsychotics\wedge parkinsonism\\
\,\,\,\,\,\,\,\,\,\,\,\,\,\,\,\,\,\,\,\,\,\,\,\,\,\,\,\,\,\,\,\,\,\,\,\,\,\,\,\,\,\,\,\,\,\,\,\,\,\,\,\,\,\,\,\,\,\,\,\,\,\,\,\,\rightarrow display(parkinsonism)
\end{array}$

$\begin{array}{l}
\mathcal{R}^{u1}\left(R_{D6},Lewy\,\,Body\right)=antipsychotics\wedge\neg parkinsonism\\
\,\,\,\,\,\,\,\,\,\,\,\,\,\,\,\,\,\,\,\,\,\,\,\,\,\,\,\,\,\,\,\,\,\,\,\,\,\,\,\,\,\,\,\,\,\,\,\,\,\,\,\,\,\,\,\,\,\,\,\rightarrow display(Lewy\,\,Body)
\end{array}$

$\begin{array}{l}
\mathcal{R}^{u2}\left(R_{D6},Lewy\,\,Body\right)=antipsychotics\wedge Lewy\,\,Body\\
\,\,\,\,\,\,\,\,\,\,\,\,\,\,\,\,\,\,\,\,\,\,\,\,\,\,\,\,\,\,\,\,\,\,\,\,\,\,\,\,\,\,\,\,\,\,\,\,\,\,\,\,\,\,\,\,\,\,\,\rightarrow display(Lewy\,\,Body)
\end{array}$

For START rule D2, if fibre is already prescribed, no clinical condition
is displayed. Otherwise, only constipation is displayed. When constipation
is checked, diverticulosis is displayed.

\subsection{Formalization of the ordering problem}

Identifying the best global strategy for asking clinical conditions
can be formalized as an ordering problem: finding the optimal strict
total order $\prec$ between the clinical conditions in $C$ that
minimizes the number of distinct clinical conditions displayed in
the questionnaire when no clinical condition are entered yet, \emph{i.e.}
finding $\prec$ that minimizes:

\noindent $\begin{array}{l}
\Big|\Big\{ x\in C\mid\exists R_{j}^{d}\in R^{d}\,\,such\,\,as\,\,R_{j}^{d}=conditions\rightarrow display(x)\\
\,\,\,\,\,\,\,\,\,\,\,\,\,\,\,\,\,\,\,\,\,\,\,\,\,\,\,\,\,\,\,\,\,\,\,\,\,\,\,\,\,\,\,\,\,\,\,\,\,\,\,\,\,\,\,\,\,\,\,\,\,\,\,\,\,\,\,and\,\,conditions\,\,are\,\,satisfied\Big\}\Big|
\end{array}$

\subsection{Proof of NP-hardness}

The ordering problem can be proved to be NP-hard, by reducing it to
the Generalized Traveling Salesman Problem (GTSP). The Traveling Salesman
Problem (TSP) consists in finding the shortest travel that passes
through a given set of towns. The GTSP is similar, but also considers
a set of areas, each town being located in a given area, and the travel
must pass through one town of each area (instead of all towns). Both
TSP and GTSP are NP-hard.

For the sake of simplicity, we will restrict the proof to simpler
display rules of the form $R'=\bigwedge\left\{ c\in C^{p}\right\} \rightarrow A=\left(C^{p},\emptyset,\emptyset,\emptyset,\emptyset,A\right)$,
and we will consider that all patient conditions are false. With such
rules, the order $\prec$ can be built progressively, from left to
right. An order $\prec$ under construction is of the form $C_{a}\prec C_{b}\prec...\prec C_{\alpha}\simeq C_{\beta}\simeq...$.
In such order, conditions $C_{ord}=\left\{ C_{a},C_{b},...\right\} $
have been ordered and other conditions $C_{unord}=\left\{ C_{\alpha},C_{\beta},...\right\} $
have not yet been ordered, with $\forall c\in C_{ord}\forall c'\in C_{unord},c\prec c'$.
In the definition of $\mathcal{R}^{p}(R_{i},x)$, if we consider a
rule of the form $R'$, we can see that it will generate display rules
of the form: $True\rightarrow display(C_{a})$, $C_{a}\rightarrow display(C_{b})$,
$C_{a}\wedge C_{b}\rightarrow display(C_{c})$,... for the order $C_{a}\prec C_{b}\prec C_{c}\prec...$.
If we assume that, in the initial form, all clinical conditions are
false, the display rules $\mathcal{R}^{p}(R_{i},x)$ will lead to
the display of a single condition, the first one according to the
order $\prec$ in $C^{p}$. Let us note $C_{r}^{p}$ the $C^{p}$
component of the rule $r$. We search the order $\prec$ that minimizes
the number of clinical conditions displayed in the initial form, \emph{i.e.}:

\noindent $\left|\left\{ x\mid\exists r\in R^{d}\,\,such\,\,that\,\,x\in C_{r}^{p}\,\,and\,\,\forall y\in C_{r}^{p}\,\,with\,\,y\neq x,x\prec y\right\} \right|$

In the GTSP reformulation of the problem, the distance corresponds
to the number of clinical conditions displayed in the form. We cannot
represent each clinical condition by a town. In fact, in GTSP, when
adding a new town in the travel, the cost (\emph{i.e.} the distance)
depends only on the current town and the new town added; on the contrary,
in our problem, when adding a new clinical condition $x$ to the right
of the order $\prec$ , the cost depends on all clinical conditions
already present in the order $\prec$ under construction, and not
only on the last one. The cost of the addition is 1 (\emph{i.e.},
one new condition to display in the form) if there is at least one
rule that includes $x$ in its conditions $C^{p}$ and that includes
no other condition already present in the order $\prec$ under construction;
otherwise it is 0 (no new condition to display).

Consequently, we considered a town as being the subset of the clinical
conditions currently included in the order $\prec$ under construction
(\emph{i.e.} $C_{ord}$). Thus, the ordering problem can be rewritten
as a GTSP by considering a set of $2^{n}$ towns $T=\left\{ x\subseteq C\right\} $,
and a set of $n+1$ areas $A=\left\{ A_{0},A_{1},...,A_{n}\right\} $,
each town $t$ belonging to the area $A_{\left|t\right|}$, \emph{i.e.}
$A_{k}=\left\{ t\in T\mid\left|t\right|=k\right\} $.

The optimal strict total order can be deduced by ordering the clinical
conditions in their order of appearance in the town sets. For example,
if the solution found is $\left(\left\{ \right\} ,\left\{ C_{1}\right\} ,\left\{ C_{1},C_{3}\right\} ,\left\{ C_{1},C_{3},C_{2}\right\} \right)$,
then the optimal total order is $C_{1}\prec C_{3}\prec C_{2}$.

The asymmetric distance matrix $M$ of the GTSP is defined as follows:%
\begin{comment}
\[
M(i\in T,j\in T)=\begin{cases}
0 & if\,\,i=C\wedge j=\emptyset\\
+\infty & if\,\,\left|i\right|\neq\left|j\right|+1\\
1 & \exists r\in R\mid j\setminus i\subseteq C_{r}^{p}\wedge i\cap C_{r}^{p}=\emptyset\\
0 & otherwise
\end{cases}
\]
\end{comment}
\[
M(i\in T,j\in T)=\begin{cases}
0 & if\,\,i\in A_{n}\wedge j\in A_{0}\\
+\infty & if\,\,\neg\left(i\in A_{k}\wedge j\in A_{k+1}\right)\\
1 & \exists r\in R\mid j\setminus i\subseteq C_{r}^{p}\wedge i\cap C_{r}^{p}=\emptyset\\
0 & otherwise
\end{cases}
\]

The first condition gives a distance of 0 for closing the loop of
the travel.

The second condition gives a distance of $+\infty$ when the salesman
travels from a town in the area $A_{k}$ to a town that is not in
area $A_{k+1}$, in order to force to visit all areas in order.

The third condition gives a distance of 1 if the salesman travels
to a town that adds a new clinical condition that is present in a
rule for which no other condition is already present in the departure
town.

The distance is 0 otherwise.

\subsection{Solving the ordering problem}

NP-hard problems can only be solved by testing all possible solutions,
or by using heuristic algorithms that give a good solution, but not
necessarily the best one. In theory, the optimal clinical conditions
order is patient-dependent: in particular, rules having non-clinical
conditions only impact patients having these conditions.

We considered two options for solving the ordering problem: (1) a
simple heuristic that sorts clinical conditions in decreasing order
of their number of occurrences in the rules, producing a global, patient-independent,
order, and (2) the Artificial Feeding Birds (AFB) metaheuristic \citep{Lamy2018_3},
for computing a near optimal, patient-specific order.

\subsection{User interface design}

Each clinical condition is displayed as a checkbox in the user interface.
Checkboxes are grouped in 13 general categories (\emph{e.g.} cardiology,
digestive system, \emph{etc.}) to facilitate the search of a particular
condition. We previously designed and used these 13 categories and
their associated colors in a work focused on adverse drug events \citep{Lamy2021}.

Some conditions are associated with several codes in the terminology,
\emph{e.g.} ``diabetes'' can be associated with ICD10 (International
Classification of Diseases, 10\textsuperscript{th} revision) codes
E10 (insulin-dependent diabetes mellitus), E11 (non-insulin-dependent
diabetes mellitus), \emph{etc}. In that case, when the box is checked,
a drop-down combo box allows the user to select the appropriate term.
By default, the most general term is selected, \emph{e.g.} E14 (unspecified
diabetes mellitus).

In rare cases, a clinical condition may be more general than another
one, \emph{i.e.} related together with ``is a'' relation (\emph{e.g.}
``diabetes'' is more general than ``type 2 diabetes''). In that
case, only the most general condition is displayed (\emph{e.g.} ``diabetes''),
and the user may select the desired code in the drop-down combo box
(including the code for the more specific condition).

When new conditions appear during user interaction, following the
selection of a given condition, the new conditions are highlighted
in yellow and marked with a red star.

\subsection{Application to STOPP/START v2}

STOPP/START v2 is a clinical guideline for medication reviews \citep{O'Mahony2015}.
It includes 114 recommendations, with both recommendations for stopping
current drug treatment (STOPP) and for starting new prescriptions
(START). 

Our implementation was written in Python 3 with the Owlready 2 ontology-oriented
programming module \citep{Lamy2017_5} for Python. Clinical rules
were expressed in a high-level language that is automatically translated
into SPARQL queries. Clinical conditions were coded using ICD10. For
more details on the implementation of STOPP/START, please refer to
the coded algorithms proposed by CJA Huibers \emph{et al.} \citep{Huibers2019}
and our previous publication for the global architecture of the system
\citep{Mouazer2022}.

\subsection{Evaluation methods}

In a first experiment, the proposed method was tested on 10 realistic
clinical cases designed by a GP (HF). We evaluated both the number
of clinical conditions displayed and the time required for executing
the display rules. The method was executed on a modern laptop computer
(processor: Intel Core i7-10510U CPU, 1.80 GHz).

In a second experiment, we organized focus group sessions for presenting
ABiMed, our CDSS for medication reviews, to clinicians (both GPs and
pharmacists). A prototype of our CDSS was shown, various user interfaces
focused on polypharmacy management were presented during the focus
groups, including the proposed adaptive questionnaire and the implementation
of the STOPP/START v2 guideline. Then, a clinical case was given and
solved together, using the support of the CDSS. We asked the opinion
of the clinicians, and in particular whether the fact that checkboxes
may appear or disappear during the data entry would be disturbing,
or not. Focus groups were tape-recorded and then transcribed.

\section{\label{sec:Results}Results}

\subsection{Application to STOPP/START v2}

The proposed methods were applied to STOPP/START v2. Three recommendations
were considered as too general for implementation (STOPP A1, A2 and
A3). The other 111 recommendations were translated into 124 clinical
rules (a few recommendations were split in two rules). The implementation
of STOPP/START v2 considered a total of 73 distinct clinical conditions
(55 for STOPP rules and 30 for START, 12 conditions belonging to both).
Using the proposed methods, 197 display rules were generated (136
for STOPP rules and 61 for START).

\begin{figure*}
\noindent \begin{centering}
\includegraphics[width=1\textwidth]{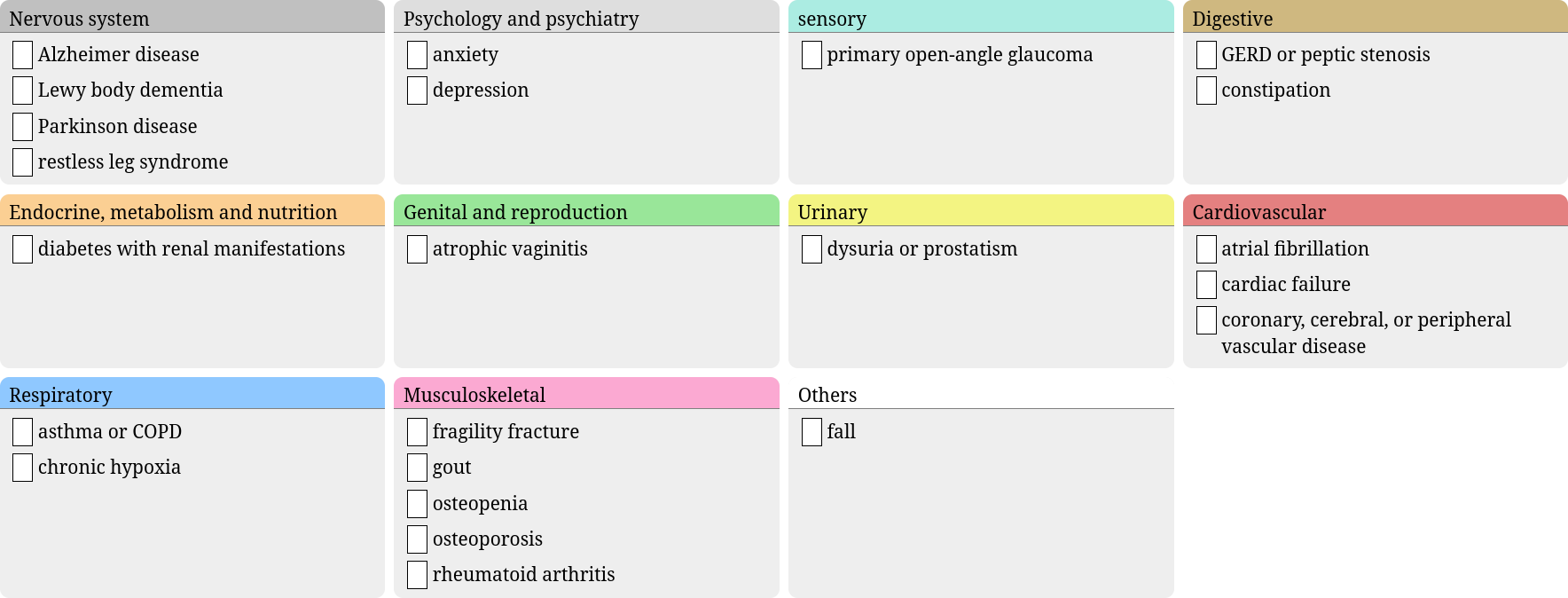}
\par\end{centering}
\caption{\label{fig:questionnaire0}The adaptive questionnaire for a patient
having an empty drug order.}
\end{figure*}

\begin{figure*}
\noindent \begin{centering}
\includegraphics[width=1\textwidth]{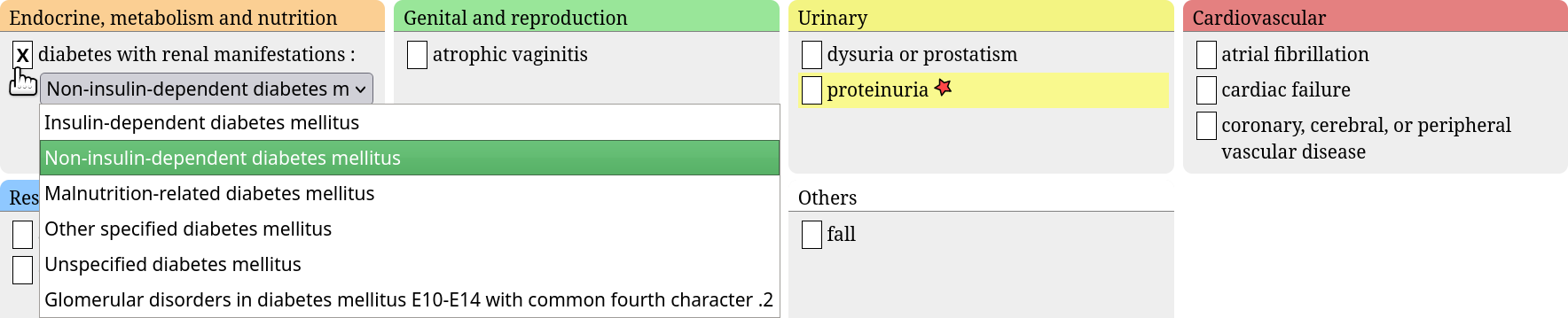}
\par\end{centering}
\caption{\label{fig:questionnaire1}Excerpt of the adaptive questionnaire of
Figure \ref{fig:questionnaire0}, after the user checked the condition
``diabetes with renal manifestation''. Notice the drop-down combo
box for choosing the appropriate ICD10 term, and the new condition
``proteinuria'' that appeared on the right.}
\end{figure*}

\begin{table*}[tp]
\begin{centering}
\begin{tabular}{cccccccccccc}
\hline 
 &  &  & \textbf{Execution} &  & \multicolumn{3}{c}{\textbf{Conditions displayed}} &  & \multicolumn{3}{c}{\textbf{Rules triggered}}\tabularnewline
\textbf{Case} & \textbf{\# drugs} &  & \textbf{time} &  & STOPP (55) & START (30) & Total\textsuperscript{1} (73) &  & STOPP (77) & START (34) & Total (111)\tabularnewline
\hline 
\#1 & 11 &  & 0.075 s &  & 11 & 21 & 28 &  & 5 & 3 & 8\tabularnewline
\#2 & 12 &  & 0.065 s &  & 6 & 22 & 27 &  & 0 & 6 & 6\tabularnewline
\#3 & 15 &  & 0.061 s &  & 5 & 17 & 20 &  & 4 & 4 & 8\tabularnewline
\#4 & 9 &  & 0.058 s &  & 7 & 18 & 24 &  & 3 & 5 & 8\tabularnewline
\#5 & 6 &  & 0.057 s &  & 4 & 20 & 24 &  & 2 & 3 & 5\tabularnewline
\#6 & 11 &  & 0.064 s &  & 16 & 20 & 29 &  & 8 & 6 & 14\tabularnewline
\#7 & 13 &  & 0.072 s &  & 6 & 20 & 25 &  & 3 & 4 & 7\tabularnewline
\#8 & 15 &  & 0.066 s &  & 10 & 19 & 29 &  & 3 & 3 & 6\tabularnewline
\#9 & 10 &  & 0.061 s &  & 13 & 22 & 34 &  & 4 & 2 & 6\tabularnewline
\#10 & 13 &  & 0.060 s &  & 8 & 20 & 27 &  & 1 & 4 & 5\tabularnewline
\hline 
\textbf{Mean} & 11.5 &  & 0.064 s &  & 8.6 & 19.9 & 26.7 &  & 3.3 & 4.0 & 7.3\tabularnewline
\textbf{(\%)} &  &  &  &  & (15.6\%) & (66.3\%) & (36.6\%) &  &  &  & \tabularnewline
\hline 
\end{tabular}
\par\end{centering}
\caption{\label{tab:Results_on_10_cases}Execution time for display rules,
number of conditions displayed and number of rules triggered for 10
representative clinical cases of elderly patients with polypharmacy.
\protect\textsuperscript{1}: the total of conditions displayed may
be less than the sum of STOPP and START (\emph{e.g.} case \#1) , because
some conditions are common.}
\end{table*}

\begin{comment}
, or sometime more than the sum (\emph{e.g.} case \#8), because the
strategy and the order of priority between conditions differ
\end{comment}

\subsection{User interface}

Figure \ref{fig:questionnaire0} shows the adaptive questionnaire
we designed, for a patient having an empty drug order. There are 23
conditions displayed, grouped in anatomical categories. Each category
is represented by a panel with a colored title bar, and each condition
is represented by a checkbox. Panels are organized on 4 columns, in
order to display all conditions on a single screen, without having
to scroll up and down.

As a matter of comparison, supplementary file \#1 shows the entire
questionnaire, if it was not adaptive, thus displaying all clinical
conditions.

Figure \ref{fig:questionnaire1} shows the same questionnaire after
the user checked the ``diabetes with renal manifestation'' condition.
In addition to the check in the checkbox, the questionnaire was adapted
on two points. First, a drop-down combo box is now displayed for choosing
the appropriate ICD10 term (since there are several ICD10 terms that
correspond to diabetes with renal manifestation). Second, a new condition,
``proteinuria'', appeared on the right. This condition was not necessary
for applying STOPP/START rules before checking ``diabetes with renal
manifestation'', but becomes necessary after. The new condition is
temporarily highlighted with a yellow background, to attract user
attention. It also has a red star after its label. The red star is
more subtle than the yellow background, but remains permanently.

Similarly, conditions may disappear during user interaction, \emph{e.g.}
the ``proteinuria'' condition would disappear if the user unchecks
the ``diabetes with renal manifestation'' condition.

\subsection{Evaluation results}

Table \ref{tab:Results_on_10_cases} shows the results obtained when
applying the proposed methods to 10 realistic clinical cases of old
patients with polypharmacy, using the simple heuristics described
in the methods section for finding a global order of the clinical
conditions. The execution time remains lower than 0.1 second on a
modern computer, which is compatible with a clinical use. When using
the AFB metaheuristic to compute a patient-specific clinical conditions
order, the algorithm quickly converged, and we obtained exactly the
same results in terms of the number of conditions displayed, but the
execution time was longer (about 0.4 second). We thus keep the simple
heuristic.

Out of the 73 clinical conditions required for STOPP/START v2, only
26.7 conditions are displayed on average, which is 36.6\% (almost
a two-third reduction). This represents an important reduction in
the number of conditions to enter or review manually. More specifically,
the reduction is much more important for STOP (only 15.6\% of the
conditions are displayed), than for START (66.3\% of the conditions
are displayed). This was expected, because all STOP recommendations
have a drug condition that must be present (\emph{i.e.} $D^{p}\neq\emptyset$),
thus allowing displaying the clinical conditions of the recommendation
only if the drug is present. On the contrary, START recommendations
may have no such drug condition. Nevertheless, we observe that the
proposed method still substantially reduces the number of conditions
displayed for START.

\begin{comment}
c’était plutôt vraiment simple d’utilisation, simple visuel surtout
et que vraiment on avait les données juste dont on avait besoin en
fait par rapport à la dynamique sur l’ interrogatoire, vraiment tout
ce qui était un peu polluant pour nous au niveau de l’ écran qui n’
apparaissait pas ou n’ apparaissait .. {*}{*} inaudible {*}{*} dans
la même mesure en tout cas. Donc sur ça {*}{*} inaudible {*}{*} que
ce soit pour le patient ou notre utilisation à nous à la pharmacie,
après là où il va falloir qu’on regarde 

C’est au niveau de la temporalité{*}{*} inaudible {*}{*}

Combien de temps ça prend, comment faire et là où il va y avoir le
plus gros enjeu,

je crois qu’ on le sait tous, c’ est dans l’ inclusion des patients
si tout le monde joue le jeu, à un certain moment sur une certaine
période

{*}{*} inaudible {*}{*}

Qu’on profite entièrement des ressources et de l’utilité que ça peut
avoir.
\end{comment}

We organized two focus group sessions, with a total of 16 clinicians
(8 GPs and 8 pharmacists). The first session was performed in a rural
environment, with 4 pharmacists (2 males, 2 females) and 4 GPs (4
males). All clinicians were young (30-40 years old). Seven clinicians
were young (30-40) and one was older (> 65). The second session was
performed in a city environment, with 4 pharmacists (3 males, 1 females)
and 4 GPs (2 males, 2 females). Four were young (30-40) and four were
older (50-65).

Clinicians understood easily how the adaptive questionnaire was working,
and its interest for reducing and simplifying data entry. They found
it easy to use and potentially useful for gaining time. Clinicians
said that ``it was pretty easy to use'', ``visually simple'' and
that ``really we had just the data we needed in fact in the adaptive
questionnaire, really everything that was a bit polluting for us at
the level of the screen did not appear''. Clinicians appreciated
having automatic data extraction from the EHR when possible, but also
the possibility to manually review and complement the data in the
adaptive questionnaire, or enter it from scratch when data extraction
is not possible (\emph{e.g.} if the GP uses an EHR that is not compatible,
or does not agree on data transfers). More generally, all clinicians
agreed that the CDSS will be useful to them and that the interfaces
were appropriate.

\section{\label{sec:Discussion}Discussion}

In this paper, we designed an adaptive questionnaire for facilitating
patient condition entry in clinical decision support systems, with
an application to polypharmacy and STOPP/START v2. We showed that
this approach is able to reduce by almost two thirds the number of
clinical conditions asked to the clinician, while having a very fast
execution time.

In the presented work, we distinguished clinical conditions, typically
disorders that the clinician has to enter, and non-clinical conditions,
such as drug prescriptions and lab test results, that are usually
available as structured data in EHR. However, if prescriptions or
lab test results would not be available, the method can easily be
adapted by treating them as we treated clinical conditions.

We compared two methods for ordering the clinical conditions: a simple
heuristic and a metaheuristic (AFB). Both yielded the same results
in terms of the number of conditions displayed, despite the metaheuristic
is more sophisticated and allowed a patient-specific order. This can
probably be explained by the fact that the ordering problem in theoretically
NP-hard, but in practice, many clinical conditions are unrelated and
independent from each other, at least in the STOPP/START v2 guideline.
However, this might not be the case with other guidelines, thus having
a more sophisticated option for solving the ordering problem remains
interesting.

We opted for an unordered adaptive questionnaire, while most existing
approaches are ordered (as seen in section \ref{sec:Related-works}).
This choice was motivated by the fact that clinicians usually expect
clinical conditions to be classified by anatomy (\emph{e.g.} cardiac,
renal...) and/or by etiology (\emph{e.g.} infectious diseases), as
they are in medical terminologies such as ICD10. On the contrary,
in an ordered adaptive questionnaire, clinical conditions would have
been ordered in the optimal order of entry, which is counterintuitive
for clinicians.

We applied the proposed method to STOPP/START v2, a guideline for
managing polypharmacy. This guideline is particularly favorable to
the use of adaptive questionnaires, because all STOPP rules have at
least one non-clinical condition (\emph{i.e.} a drug prescription).
If the method was applied to a guideline aimed at the treatment of
a given disorder, such as type 2 diabetes or hypertension, the reduction
of the number of conditions displayed is expected to be lower. However,
we've seen in the results section that the adaptive questionnaire
still reduce by about one third the number of conditions displayed
for START rules. For other guidelines, the reduction is expected to
be similar.

In the literature, the proposed CDSS for polypharmacy rely on the
manual entry of clinical conditions from a medical terminology or
a thesaurus (examples of such systems are PIM-Check \citep{Desnoyer2017},
KALIS \citep{Shoshi2017} and PRIMA-eDS \citep{Rieckert2018}) or
on the automatic extraction of clinical conditions from the EHR (\emph{e.g.}
in STRIPA \citep{Meulendijk2015_2}, TRIM \citep{Niehoff2016} and
MedSafer \citep{McDonald2019}). For manual data entry, PIM-Check
uses a flat list of about 150 terms. PRIMA-eDS also uses a flat list.
In the PRIMA-eDS evaluation, the users explicitly reported that the
manual data entry was ``inconvenient and time-consuming'', preventing
the use of the tool in daily practice \citep{Rieckert2018}. The other
studies relying on EHR data extraction, however, the connection to
EHR is usually difficult to achieve, especially for community pharmacists
: EHR are located at hospital or at the GP office, and not available
to community pharmacists. In practice, EHR connection was achieved
only in hospital or for GPs, or for pharmacists in very specific and
centralized situations (namely, the Veterans Affairs in US, which
operates centers integrating GPs and pharmacists, for TRIM \citep{Niehoff2016}).
But CDSS for polypharmacy are of particular use for community pharmacists
when they perform medication reviews. As a consequence, it is very
important to facilitate the manual data entry in polypharmacy CDSS,
since the expected users usually do not have access to EHR.

The main limitation of this study is that, in most contexts other
than polypharmacy, CDSS are used by physicians and not by pharmacists,
and thus the users have access to the patient data in the EHR. However,
the proposed adaptive questionnaire remains interesting as a tool
for verifying at a glance the absence of error and missing data in
the extracted patient data, before running the CDSS, or when the CDSS
require very specific or trivial data that is unlikely to be present
in the EHR. We encountered such a situation recently when designing
a CDSS for the diagnosis and management of Covid-19 patients: the
CDSS required many symptoms (\emph{e.g.} cough, rhinorrhea,...) which
are usually not entered in the EHR \citep{Lamy2023}.

The perspective of this work includes the evaluation of the STOPP/START
v2 implementation in clinical settings on real patients, and the application
of the adaptive questionnaire to other clinical practice guidelines.
The method may also be extended to more complex rule format, for instance
rules of the form ``if at least \emph{x} conditions are present in
a given list of \emph{y} conditions, then...''. Finally, adaptive
questionnaires might also be of interest for collecting patient-reported
outcomes and more generally for data entry by the patient itself.

\section{\label{sec:Conclusion}Conclusion}

In conclusion, we proposed a method based on adaptive questionnaires
for facilitating data entry in clinical decision support systems,
and we applied the proposed methods to STOPP/START v2, a clinical
practice guideline for the management of polypharmacy. The method
considers a guideline implemented as a rule-based system, and proposes
formula for translating the rules formalized from the guideline into
rules determining which clinical conditions are mandatory to display
to the clinician. Tests on STOPP/START v2 showed that this method
can reduce by almost two thirds the size of the questionnaire.

\section*{Supplementary files}

Supplementary file \#1: screenshot of the entire questionnaire, with
the ``adaptive'' functionality disabled. It displays all clinical
conditions for STOPP/START v2.

\section*{Acknowledgments}

This work was funded by the French Research Agency (ANR) through the
ABiMed project {[}Grant No. \mbox{ANR-20-CE19-0017}{]}.

\bibliographystyle{elsarticle-num}
\bibliography{biblio_ama}

\end{document}